\documentclass[journal]{IEEEtran} 

\usepackage{amsmath,amssymb,amsfonts, bm,mathdots,amsthm}

\usepackage{graphicx}
\usepackage{float}
\graphicspath{{Pictures/}{../jpeg/}} 
\usepackage[table]{xcolor}

\definecolor{lightbluebg}{RGB}{220,235,247}

\usepackage{array}
\usepackage{multirow}
\usepackage{booktabs}
\usepackage{longtable}
\usepackage{adjustbox}

\usepackage{algorithm}
\usepackage{algpseudocode} 

\ifCLASSOPTIONcompsoc
  \usepackage[caption=false,font=normalsize,labelfont=sf,textfont=sf]{subfig}
\else
  \usepackage[caption=false,font=footnotesize]{subfig}
\fi

\usepackage{xcolor,soul}
\definecolor{light}{rgb}{0.5, 0.5, 0.5}

\usepackage{cite}
\usepackage{hyperref}

\usepackage{gensymb}
\usepackage{textcomp}
\usepackage{pifont}
\usepackage{enumitem}
\usepackage{varioref}
\usepackage{latexsym}

\newcommand{\cmark}{\ding{51}}%
\newcommand{\xmark}{\ding{55}}%

\def\BibTeX{{\rm B\kern-.05em{\sc i\kern-.025em b}\kern-.08em
    T\kern-.1667em\lower.7ex\hbox{E}\kern-.125emX}}

\begin{document}
\title{A Unified BERT-CNN-BiLSTM Framework for Simultaneous Headline Classification and Sentiment Analysis of Bangla News}


\author{%
Mirza Raquib, 
Munazer Montasir Akash, 
Tawhid Ahmed, 
Saydul Akbar Murad, 
Farida Siddiqi Prity,  
Asif Pervez Polok,
Mohammad Amzad Hossain, 
Nick Rahimi$^{**}$%
\thanks{$^{**}$Corresponding author.}%
\thanks{Mirza Raquib: Department of CCE, International Islamic University Chittagong, Chattogram, Bangladesh; Department of ICE, Noakhali Science and Technology University, Noakhali, Bangladesh. Email: mirzaraquib@iiuc.ac.bd}%
\thanks{Munazer Montasir Akash: Department of CSE, Banladesh University of Engineering and Technology, Dhaka, Bangladesh; Dept. of CSE, IIUC, Chattogram, Bangladesh. Email: 0423052104@grad.cse.buet.ac.bd}%
\thanks{Tawhid Ahmed: Department of CSE, Banladesh University of Engineering and Technology, Dhaka, Bangladesh. Email: tawhidahmed98@gmail.com}%
\thanks{Saydul Akbar Murad: School of Business and Technology 
Emporia State University, KS, USA Email: smurad@emporia.edu}%
\thanks{Farida Siddiqi Prity: Department of CSE, Netrokona University, Netrokona, Bangladesh. Email: prity@neu.ac.bd}%
\thanks{Asif Pervez Polok: mPower Social Enterprise.
Email: asifpolok.research@gmail.com}%
\thanks{Mohammad Amzad Hossain: Department of ICE, Noakhali Science and Technology University, Noakhali-3814, Bangladesh. Email: amzad@nstu.edu.bd}%
\thanks{Nick Rahimi: School of Computing Sciences and Computer Engineering, University of Southern Mississippi, Hattiesburg, MS, USA. Email: nick.rahimi@usm.edu}%
}

\maketitle

\begin{abstract}
\noindent{Automatic analysis of Bangla news headlines has become increasingly important because of the rapid growth of online news platforms. However, effectively navigating the vast amount of news content from different newspapers and online news portals can be challenging. Joint headline classification and sentiment analysis enable automatic identification of both topical categories and emotional polarity. This helps us quickly understand the emotional tone of the news. This paper proposes a unified BERT-CNN-BiLSTM framework for joint Bangla headline classification and sentiment analysis using the publicly available BAN-ABSA dataset, which contains 9,014 Bangla news headlines. The dataset is highly imbalanced, leading to overfitting and poor generalization. This work addresses these challenges by applying undersampling, oversampling, and back-translation augmentation to the training set. The validation and test sets have not been changed to ensure fair and unbiased evaluation. The proposed framework has been tested in four different training regimes: imbalanced datasets, train-undersampled datasets, train-oversampled datasets, and train-augmented datasets. The proposed model is also compared with other pretrained deep learning models. The experimental results demonstrate that the proposed model is superior. The best headline classification accuracy was obtained on the original imbalanced dataset at 81.37\%, and the best sentiment classification accuracy on the train-augmented dataset at 66.40\%. The robustness and generalizability of the proposed framework are further illustrated through five-fold cross-validation, external dataset evaluation, and LIME-based interpretability analysis.
} 
\end{abstract}
\begin{IEEEkeywords}
Artificial Intelligence, Natural Language Processing,
Headline Classification, Sentiment Analysis
\end{IEEEkeywords}

\section{Introduction}
\label{sec:1}
Automatic text analysis has become an important research area due to the rapid growth of digital content and online news platforms. It is also widely applied in social media analysis, online journalism, and public opinion mining. News classification groups headlines into various categories that users can use to easily locate information of interest. It also helps news organizations to automatically process and filter huge amounts of news articles. Furthermore, sentiment analysis can reveal the sentiment expressed in a headline and help understand the hidden message. These technologies also offer valuable information for media analysts and policymakers to understand public opinion. However, Bangla remains a low-resource language in natural language processing (NLP) compared with English and other high-resource languages. Newspapers remain a main source of information, and the headlines give a brief summary of the contents of the news. In addition to the choice of headline topic, it is equally important to comprehend the sentiment of the headline that will affect how readers interpret the news. Thus, modeling for the classification of Bangla headlines and sentiment analysis of sentences has become an important research challenge.

One of the basic tasks of natural language processing (NLP) is text categorization. It is a supervised learning problem of categorizing text documents based on their linguistic and contextual information. The classic machine learning (ML) approaches, including Support Vector Machine (SVM), Naïve Bayes (NB), and Random Forest (RF), have been extensively employed to classify texts. These techniques are primarily feature-driven and generally lack the ability to encode the semantic content and sequential structure of natural language. Text classification has been greatly advanced by deep learning (DL) models, which learn feature representations directly from the data. Models such as Convolutional Neural Networks (CNNs), Recurrent Neural Networks (RNNs), Long Short-Term Memory (LSTMs), and Gated Recurrent Units (GRUs) have achieved better performance than traditional ML methods across many NLP tasks. Although these developments have brought great improvement to Bangla, it still has limited pre-trained models, a small number of annotated datasets, and a complex morphology. These challenges hinder accurate headline identification and sentiment determination. This study proposed a powerful and comprehensive hybrid model based on the news headline classification and sentiment analysis of Bangla language news headlines to address these challenges.

The performance of text classification has been enhanced by many machine learning (ML), deep learning (DL), and hybrid models. Traditional ML techniques \cite{mustafa2024review} \cite{palanivinayagam2023twenty} \cite{hassan2022analytics} \cite{garrido2023comparing} are efficient on small datasets and require minimal computational resources. But they are highly dependent on manually designed features and are unable to capture the semantics of natural language and the meaning of context. Deep learning models like CNN, BiLSTM, and GRU \cite{roy2023bengali}\cite{salehin2021comparative}\cite{alam2020bangla}\cite{emon2019deep} are able to self-learn feature representations and have proven to work on many text classification tasks with great performance. But these models tend to overfit and lack generalization, particularly on small and imbalanced Bangla datasets. In addition, hybrid models such as CNN-LSTM, CNN-BiGRU, and BERT-based classifiers have enhanced classification performance by combining contextual and sequential feature learning capabilities, as demonstrated in \cite{hasib2023mcnn}, \cite{saha2022sentiment}, \cite{alam2022bengali}, \cite{mahmud2024enhancing}, and \cite{hassan2024analyzing}. Although significant progress has been made, most current work focuses only on moderate class imbalance and frequently rescales the entire dataset to evaluate performance, potentially skewing the results. Furthermore, most previous work has focused on headline classification or sentiment analysis only. A few studies explore both tasks within a single approach. These restrictions inspire the proposed BERT-CNN-BiLSTM framework for Bangla headline classification and sentiment analysis.

The BAN-ABSA\cite{Ahmed2020BANABSA} dataset is highly imbalanced, thus posing the risk of overfitting and under-generalization. This paper proposes a unified BERT-CNN-BiLSTM framework for joint Bangla headline classification and sentiment analysis to address this challenge. The proposed method uses only oversampling and back-translation for the training set, leaving the other sets (validation and test) untouched to ensure a fair and unbiased evaluation. Four different training settings are explored: original imbalanced data, train-undersampled, train-oversampled, and train-augmented. The different training settings investigated to study the effect of different balancing strategies are train-undersampled, train-oversampled, train-augmented data, and the original imbalanced data. Additionally, the framework incorporates contextual embeddings, local feature extraction, sequential learning, and regularization techniques for better feature representation and to prevent overfitting. 5-fold cross-validation and LIME interpretability analysis were employed to check the robustness and reliability of the models, with extensive experiments performed using Accuracy, Precision, Recall, and F1-score. The proposed framework sets a high reference for the BAN-ABSA dataset, and it is an effective method for dealing with class imbalance in Bangla text classification without inducing evaluation bias.

\subsection{Major Contributions of This Work}
The major technical contributions of this research are summarized as follows:
\begin{itemize}
    \item We have proposed a unique BERT-CNN-BiLSTM framework for the joint task of Bangla headline classification and sentiment analysis, validated through extensive ablation studies.

    \item We have solved the overfitting problem on this dataset using efficient fine-tuning and class-balancing strategies.

    \item We have achieved state-of-the-art accuracy, and generalized our model by experimenting with it on external datasets.
\end{itemize}

\subsection{Organization}
The rest of this article is organized as follows: the related works are summarized in Section \ref{sec:2}. In Section \ref{sec:3}, we describe the proposed methodology in details, which consists of the dataset collection, processing, and train validation test dataset,. In Section \ref{sec:4}, the recognition experimental results are presented with a detailed explaination on evaluation criteria of the proposed model. Finally, our conclusion is given in Section \ref{sec:5}.

\section{Literature Review}
\label{sec:2}
The rapid growth of digital media platforms has increased the importance of Bangla news headline classification and sentiment analysis. These two activities will also enable readers to grasp the content and mood of news headlines. Previous studies have mainly focused on headline classification and sentiment analysis separately. But in real life, news reading demands that readers simultaneously determine the news item's category and its sentiment. The growing number of Bangla online news portals has further increased the need for integrated analysis. Class imbalance is also another significant problem, which is likely to affect the accuracy of the classification and generalization of the models. Researchers have applied undersampling, oversampling, and data augmentation techniques to address this issue. In the next section, the existing literature on Bangla headline classification, sentiment analysis, and hybrid deep learning models is reviewed.

Even though there is a scarcity of benchmark datasets and good word embeddings, several previous works have studied automatic Bangla news headline classification. Some of the available studies on both headline and sentiment analysis are tabulated in Table \ref{tab:literature review}. An early study suggested an LSTM-based deep learning model based on 4580 Bangla news headlines collected and established a good baseline for Bangla news headline classification \cite{bhuiyan2021approach}. Later research has investigated deep learning and hybrid models, which enhance the classification accuracy. The authors, for instance, compared CNN, BiLSTM, and hybrid BiLSTM+SVM with the traditional machine learning methods using TF-IDF and Word2Vec embeddings. Their hybrid model achieved the best performance with an accuracy of 94\% and an F1-score of 94\% \cite{hossain2025banglanewsclassifier}. In another study, several classical machine learning classifiers, such as SVM, Naïve Bayes, Logistic Regression, and Random Forest, were evaluated along with the neural network model. The accuracy of the neural network model was found to be 90\%, which is the highest among all the models evaluated in this article \cite{khushbu2020neural}. The Bangla news headline classification problem is also studied based on the GRU model. The authors have mentioned that the GRU model performed best among four machine learning and three deep learning models with the highest accuracy of 84\% in all eight headline categories in the Bengali language \cite{chowdhurybengali}. In another research \cite{chowdhury2021bangla}, a CNN-LSTM model with GloVe word embeddings (word representations) was found to outperform several state-of-the-art methods and attain an accuracy of 87\%. In more recent work, NewsNet, which is a hybrid CNN-RNN architecture with BiLSTM and parameterized GRU, reported an accuracy of 94.57\% with precision, recall, and F1 scores above 94\% \cite{rana2024newsnet}. However, most of these works used static word embeddings (TF-IDF, Word2Vec, GloVe) or single-task learning problems, and yielded satisfactory results, but not high. The studies achieved satisfactory results, but most of them used static word embeddings like TF-IDF, Word2Vec, and GloVe, and most of them considered only single-task learning problems. This makes them less effective in the context of modern NLP models, transformer-based systems, because they cannot fully capture the contextual semantics and cannot model topics and sentiment in the headline together.

Some works have explored Bangla sentiment analysis using traditional ML, Deep learning,, and hybrid models. Preliminary work tested Logistic Regression (LR), SGD, SVM, Naïve Bayes (NB), Decision Tree, Random Forest (RF), and LSTM models on 4K Bangla comments with features extracted using Count Vectorizer and TF-IDF. The LSTM model outperforms with an accuracy of 84.25\% \cite{saha2021sentiment}. Later, a hybrid CNN-LSTM model with multi-word embeddings based on Word2Vec, GloVe, and CBOW, along with an embedding layer, was proposed on an annotated dataset of 6 emotions comprising 32000 Facebook comments. Howeverdue toof class imbalance, only thref emotioclasses n were consider,ed, and the hybrid modachieved theave the highest accuracy (90.49\%) and F1 score (92.83\%) compared to traditional ML and deep learning models \cite{hoq2021sentiment}. In another work, Bangla product review classification was considered as the task using TF-IDF and a machine learning algorithm, and Gaussian NB (GNB) achieved maximum accuracy of 88.23\% \cite{karmakar2022sentiment}. Bhowmik \textit{et al.} introduced their own model named Bangla Text Sentiment Score (BTSC) and combined it with various deep learning models with Word2Vec embeddings. BERT-LSTM model showed the best accuracy of 84.18\% among CNN, RNN, Bi-LSTM, HAN-LSTM, D-CAPSNET-Bi-LSTM, and BERT-LSTM models \cite{bhowmik2022sentiment}. A hybrid CNN-LSTM (CLSTM) model has been developed to classify multi-class sentiment from Facebook comments, which yielded 85.8\% accuracy and an F1 score of 0.86 on a 42,036-comment dataset in four classes of sentiment \cite{haque2023multi}. Samin \textit{et al.} evaluated the sentiment of Bangladeshi newspaper articles by applying six machine learning classifiers, and among all classifiers, Logistic Regression gave the best accuracy of 85.19\% in three-class sentiment classification \cite{samin2024sentiment}. Recently, transformer-based models have shown much better results than traditional models. When compared with traditional ML models, the BERT-based model achieved 93\% accuracy and 92\% F1-score, which are significantly higher results in Bangla sentiment analysis compared with the other models \cite{mahmud2025transformers}. While most of these studies have reported positive results, they have been based on relatively small or imbalanced datasets and have considered only binary or ternary sentiment classification. Traditional machine learning techniques rely on handcrafted features, and many hybrid approaches still suffer from overfitting and poor generalization. The limitations indicate the need for more powerful hybrid models for Bangla sentiment analysis using the transformer architecture.

Current research on Bangla news analysis has three major problems: (i) it uses static embeddings and shallow architectures, (ii) it doesn't model both headline category and sentiment polarity at the same time, and (iii) it doesn't look at class imbalance in low-resource settings enough. These gaps create the need for a single, transformer-based framework that can handle data imbalance and capture both local and global semantic dependencies. This study addresses these constraints. We have introduced a cutting-edge hybrid deep learning framework, BERT-CNN-BiLSTM, for concurrent classification of Bangla news headlines and sentiment analysis. The suggested model combines contextualized transformer embeddings (BERT), convolutional feature extraction (CNN), and bidirectional sequential modeling (BiLSTM). The architecture enables the model to learn both topical and emotional representations at the same time. This study also systematically examines three imbalance-handling strategies. The assessment of the model using the BAN-ABSA dataset, which has not been previously analyzed in this integrated framework.

\begin{table*}[htbp]
\centering
\renewcommand{\arraystretch}{1.5}
\resizebox{\textwidth}{!}{\begin{tabular}{|p{2.5cm}|c|c|c|c|c|p{2.5cm}|p{2cm}|c|c|c|c|c|c|p{4.8cm}|}
\hline
\multirow{2}{*}{\textbf{Study}}
& \multicolumn{5}{c|}{\textbf{Dataset}}
& \multirow{2}{*}{\textbf{DL Models}}
& \multirow{2}{*}{\textbf{Hybrid Models}}
& \multicolumn{6}{c|}{\textbf{Metrics}}
& \multirow{2}{*}{\textbf{Limitations}} \\
\cline{2-6} \cline{9-14}
& \rotatebox{90}{\textbf{Headline}}
& \rotatebox{90}{\textbf{Sentiment}}
& \rotatebox{90}{\textbf{Other Dataset}}
& \rotatebox{90}{\textbf{Head. Classes}}
& \rotatebox{90}{\textbf{Sent. Classes}}
&
&
& \rotatebox{90}{\textbf{Acc.}}
& \rotatebox{90}{\textbf{Pre.}}
& \rotatebox{90}{\textbf{Rec.}}
& \rotatebox{90}{\textbf{F1}}
& \rotatebox{90}{\textbf{Cross Val.}}
& \rotatebox{90}{\textbf{XAI}}
& \\
\hline

\cite{bhuiyan2021approach} & $\checkmark$ & $\times$ & $\times$ & 5 & $\times$ & LSTM & $\times$
& $\checkmark$ & $\checkmark$ & $\checkmark$ & $\checkmark$ & $\times$ & $\times$
& Small dataset, weak embeddings, no lemmatiser \\ \hline

\cite{hossain2025banglanewsclassifier} & $\checkmark$ & $\times$ & $\times$ & 8 & $\times$ & CNN, BiLSTM & BiLSTM+SVM
& $\checkmark$ & $\checkmark$ & $\checkmark$ & $\checkmark$ & $\times$ & $\times$
& Imbalanced dataset (economic vs tech) \\ \hline

\cite{khushbu2020neural} & $\checkmark$ & $\times$ & $\times$ & 10 & $\times$ & SVM, NB, LR, RF, NN & $\times$
& $\checkmark$ & $\times$ & $\times$ & $\times$ & $\times$ & $\times$
& Small dataset, limited categories \\ \hline

\cite{chowdhurybengali} & $\checkmark$ & $\times$ & $\times$ & 8 & $\times$ & GRU & $\times$
& $\checkmark$ & $\checkmark$ & $\checkmark$ & $\checkmark$ & $\times$ & $\times$
& Moderate accuracy \\ \hline

\cite{chowdhury2021bangla} & $\checkmark$ & $\times$ & $\times$ & 10 & $\times$ & LSTM, CNN, SVM, BiLSTM, ANN & CNN+LSTM
& $\checkmark$ & $\times$ & $\times$ & $\checkmark$ & $\times$ & $\times$
& No transformer embeddings (static GloVe only) \\ \hline

\cite{rana2024newsnet} & $\checkmark$ & $\times$ & $\times$ & 8 & $\times$ & BiLSTM, GRU, CNN & NewsNet (CNN+RNN)
& $\checkmark$ & $\checkmark$ & $\checkmark$ & $\checkmark$ & $\times$ & $\times$
& Single dataset only \\ \hline

\cite{sultana2025attention} & $\checkmark$ & $\times$ & $\times$ & 6 & $\times$ & CNN, BiLSTM, GRU & CNN+BiLSTM+ (Attention)
& $\checkmark$ & $\checkmark$ & $\checkmark$ & $\checkmark$ & $\times$ & $\times$
& Limited to 6 categories, moderate accuracy \\ \hline

\cite{saha2021sentiment} & \xmark & $\checkmark$ & $\times$ & $\times$ & 2 & LSTM, LR, SGD, SVM, NB, DT, RF & $\times$
& $\checkmark$ & $\checkmark$ & $\checkmark$ & $\checkmark$ & $\times$ & $\times$
& Binary only, small dataset, no lexicons \\ \hline

\cite{hoq2021sentiment} & $\times$ & $\checkmark$ & $\times$ & $\times$ & 3 & CNN, LSTM & BERT+LSTM
& $\checkmark$ & $\checkmark$ & $\checkmark$ & $\checkmark$ & $\times$ & $\times$
& Only 3 emotions used (imbalance) \\ \hline

\cite{karmakar2022sentiment} & $\times$ & $\checkmark$ & $\times$ & $\times$ & 2 & SVM, Gaussian NB, Multinomial NB & $\times$
& $\checkmark$ & $\checkmark$ & $\checkmark$ & $\checkmark$ & $\times$ & $\times$
& Shallow ML only, binary only \\ \hline

\cite{bhowmik2022sentiment} & $\times$ & $\checkmark$ & $\times$ & $\times$ & 3 & CNN, LSTM (Word2Vec, CBOW, GloVe) & CNN+LSTM
& $\checkmark$ & $\checkmark$ & $\checkmark$ & $\checkmark$ & $\times$ & $\times$
& Small, imbalanced dataset \\ \hline

\cite{haque2023multi} & $\times$ & $\checkmark$ & $\times$ & $\times$ & 4 & CNN, LSTM, BiLSTM, GRU & CNN+LSTM (CLSTM)
& $\checkmark$ & $\checkmark$ & $\checkmark$ & $\checkmark$ & $\times$ & $\times$
& Overfitting, imbalance \\ \hline

\cite{samin2024sentiment} & $\times$ & $\checkmark$ & $\times$ & $\times$ & 3 & LR, SVM, NB, RF, KNN & $\times$
& $\checkmark$ & $\checkmark$ & $\checkmark$ & $\checkmark$ & $\times$ & $\times$
& English only, no DL for Bangla \\ \hline

\cite{mahmud2025transformers} & $\times$ & $\checkmark$ & $\times$ & $\times$ & 3 & SVM, LSTM, BERT & $\times$
& $\checkmark$ & $\checkmark$ & $\checkmark$ & $\checkmark$ & $\times$ & $\times$
& Focus on bias detection \\ \hline

\textbf{Our Model} & $\checkmark$ & $\checkmark$ & $\checkmark$ & 4 & 3 & CNN, BiLSTM & \textbf{BERT-CNN-BiLSTM}
& $\checkmark$ & $\checkmark$ & $\checkmark$ & $\checkmark$ & $\checkmark$ & $\checkmark$
& The framework is restricted to two segments (i.e. headline classification and sentiment analysis), but extension to further ones (e.g. topic categorisation, emotion detection, sarcasm identification, aspect-based sentiment analysis) are needed for more generalisation. \\ \hline

\end{tabular}}
\vspace{0.1cm}
\small
\\ \textbf{Legend:} Acc = Accuracy, Pre = Precision, Rec = Recall, F1 = F1-score, Cross Val. = Cross Validation, XAI = Explainable AI,
\\ \cmark = Yes, \xmark = No
\caption{Comparison of related studies with the proposed BERT-CNN-BiLSTM hybrid model.}
\label{tab:literature review}
\end{table*}

\begin{figure*}[t] 
    \centering
    \includegraphics[width=\linewidth]{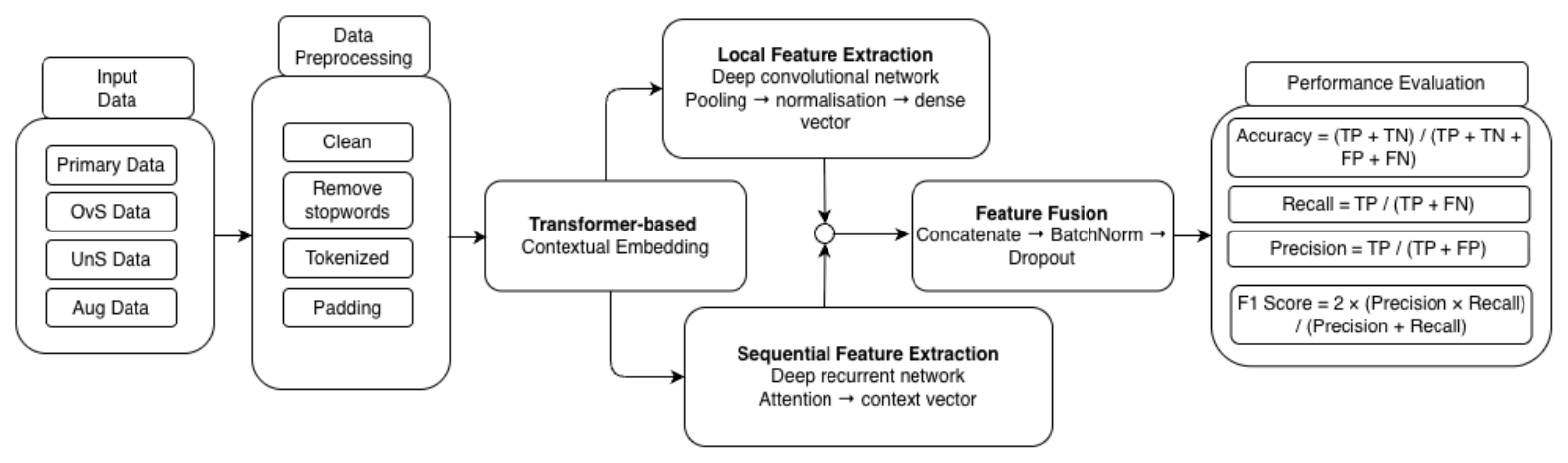}
    \caption{Flow Chart of the headline classification with sentiment analysis}
    \label{fig-workflow}
\end{figure*}

\section{Methodology} 
\label{sec:3}
The proposed model aims to classify Bangla news headlines using sentiment analysis with a single hybrid deep learning framework. It brings together the contextual language understanding power of BERT, the local feature extraction power of CNN, and the sequential learning power of BiLSTM networks. However, the hybrid design allows the model to extract contextual, semantic, and sequential information from Bangla news headlines more effectively than the individual models.

The overall workflow of the proposed framework for headline classification and headline sentiment analysis is illustrated in Figure \ref{fig-workflow}.

\subsection{Dataset Collection}
The publicly available BAN-ABSA\cite{Ahmed2020BANABSA} dataset is one of the available datasets for Bangla news headline classification and sentiment analysis used in this study. The Bangla news headlines are extracted from several online Bangla news sites, totaling 9,014. All headlines are manually categorized into headline categories and labeled with a sentiment score. This allows comparison of two text classification tasks with a common input. The headline classification task has four categories: Religion,  Politics, Sports, and Other. The sentiment classification task classifies each headline as Positive, Negative, or Neutral. One of the main challenges of the dataset is its class imbalances, which can reduce model generalization and lead to overfitting. Thus, various data-balancing techniques are studied to enhance classification accuracy under balanced and imbalanced data distributions in this study. Formally, if the dataset is denoted as
$$
D = \{(x_i, y_i^{head}, y_i^{sent}) \mid i = 1, \dots, N\}, \quad N = 9014
$$
where $x_i$ represents the headline text, $y_i^{head} \in \{1,2,3,4\}$ represents the aspect label, and $y_i^{sent} \in \{1,2,3\}$represents the polarity label, then the distribution of samples per class is imbalanced as shown above.

\subsection{Undersampling and Oversampling}
BAN-ABSA is highly imbalanced in terms of the number of headline categories or sentiment classes. This imbalance may skew the learning process in favor of the majority classes, leading to decreased classification accuracy and poor generalization. This study uses both undersampling and oversampling methods only on training data to create balanced datasets to investigate the effect of data balancing. The distribution of classes after using different sampling is shown in Table~\ref{table-tech1-tech2}.

\subsection{Data Augmented (Back Translations)}
This study uses back translation on the training set only to address class imbalance, avoiding the loss of valuable samples and the generation of exact duplicates. Back translation creates semantically equivalent texts by first translating a sentence into an intermediate language and then translating back to the original language. For this work, Bangla headlines were translated into English and back into Bangla to create new training samples while preserving meaning and class labels.

The back-translation approach produces linguistically diverse training data without losing the semantic meaning of the originals. This can also improve the model's generalization and mitigate overfitting \cite{hasib2023strategies}. The validation and test sets were deliberately kept unchanged to avoid data leakage and to provide an unbiased assessment of the proposed framework. The distribution of classes after using back-translation based augmentation is shown in Table~\ref{table-tech1-tech2}.

\begin{table*}[h!]
\centering
\resizebox{\textwidth}{!}{
\begin{tabular}{|c|c|cccc|ccc|}
\hline
\textbf{Technique} & \textbf{Dataset Type}
& \multicolumn{4}{c|}{\textbf{Aspect}}
& \multicolumn{3}{c|}{\textbf{Polarity}} \\ \hline

&  & Other & Politics & Religion & Sports
& Negative & Positive & Neutral \\ \hline

\multirow{4}{*}{\textbf{Base}}
& Initial
& 3013 & 2663 & 1755 & 1583
& 4723 & 2622 & 1669 \\ \cline{2-9}

& Train
& 2231 & 2109 & 1393 & 1259
& 3667 & 2041 & 1274 \\ \cline{2-9}

& Validation
& 395 & 283 & 184 & 160
& 531 & 293 & 200 \\ \cline{2-9}

& Test
& 387 & 271 & 178 & 164
& 524 & 288 & 195 \\ \hline

\multirow{2}{*}{\textbf{Sampling}}
& Train-Undersampling
& 1259 & 1259 & 1259 & 1259
& 1274 & 1274 & 1274 \\ \cline{2-9}

& Train-Oversampling
& 2231 & 2231 & 2231 & 2231
& 3677 & 3677 & 3677 \\ \hline

\multirow{2}{*}{\textbf{Augment(Back-Translation)}}
& Train (Before Aug.)
& 2231 & 2109 & 1393 & 1259
& 3677 & 2041 & 1274 \\ \cline{2-9}

& Train (After Aug.)
& 2231 & 2231 & 2231 & 2231
& 3677 & 3677 & 3677 \\ \hline
\end{tabular}
}
\caption{Class distribution for Aspect and Polarity}
\label{table-tech1-tech2}
\end{table*}

\subsection{Data Preprocessing}
Data preprocessing includes several key steps: data cleaning, stopword removal and stemming, tokenization, padding, word encoding and sentence representation\cite{10086954}. Each step transforms the textual input into a structure and a suitable format for the model. The following sections describe each step individually, including relevant mathematical equations.

\begin{itemize}
\item\textbf{Clean Data:} At the initial stage of imbalanced data contains different metadata, digits and some special symbols. Now removing this noise and irrelevant information from this imbalanced data to make it cleaned data ($D_cleaned$). $$D_{cleaned} = D_m \times D_d \times D_o \times D_n$$ Where $D_m$ removes meta characters, $O_d$ remove digits, $D_o$ remove special characters and $D_n$ normalises text.

\item \textbf{Stopword Removal and Stemming} The next step is the removal of stopwords, i.e, the high frequency low semantic words after cleaning the imbalanced data. In addition, stemming reduces the words to their morphological root form, which reduces further the lexical redundancy. Additionally, to ensure that queries do not comprise of meaningless tokens, we eliminate sentences with a length of less than 2 words.
$$T_{final} = {stem(t) \mid t \in T\setminus S}$$ Where T = token set, S = stopword list, and stem(t) = stem(t) applies stemming to token t.
\item \textbf{Tokenization:} The cleaned and processed text are divided into smaller linguistic units called tokens. The process of tokenization converts them into numerical representations that deep learning models can process.  BERT's tokenizer, which can handle sophisticated Bangla morphology and capture contextual nuances, was used for subword tokenization in this study. $$T_i = { token(w_j) \mid w_j \in d_i }$$ where $d_i$ is the ith headline and $w_j$ are its words.

\item \textbf{Padding:} To achieve the same input size for the proposed model, all sequences are padded or truncated to a maximum length of 300 tokens. Sequences with fewer than 300 tokens are post-padded with zeros; those longer than 300 are truncated to the first 300 tokens. This pre-processing step enables efficient batch processing and consistent model training.

\end{itemize}

\begin{figure*}[t]     \centering
    \includegraphics[width=\linewidth]{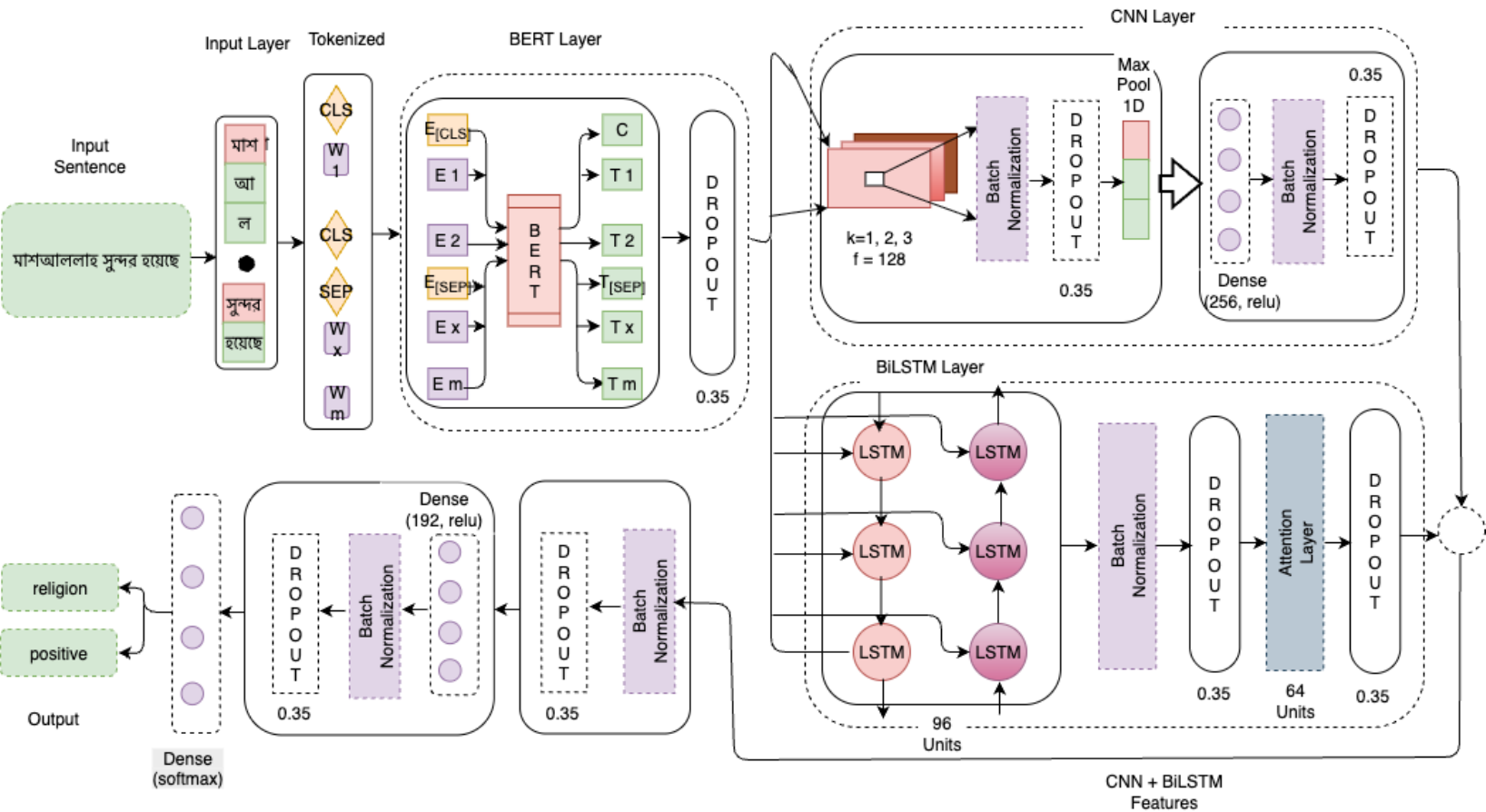}
    \caption{Model Architecture of BERT-CNN-BiLSTM}
    \label{fig-model_archi_bert-cnn-bilstm}
\end{figure*}

\begin{table}[ht]
\centering
\resizebox{\columnwidth}{!}{\begin{tabular}{|l|c|}
\hline
\textbf{Parameter} & \textbf{Value / Description} \\ \hline
Maximum sequence length ($T$) & 300 tokens \\ \hline
Hidden size ($d_{\text{bert}}$) & 768 \\ \hline
Batch size ($B$) & Depends on training configuration \\ \hline
Token IDs ($X$) & $X \in \mathbb{R}^{B \times T}$ \\ \hline
Attention masks ($M$) & $M \in \{0,1\}^{B \times T}$ (indicating valid tokens) \\ \hline
\end{tabular}}
\caption{Bert Input Configuration}
\label{tab:input_config}
\end{table}

\subsection{Proposed Model Architecture Overview}
The overall architecture of the proposed hybrid model is illustrated in Figure \ref{fig-model_archi_bert-cnn-bilstm}. The model begins with a BERT layer to generate contextualised embeddings from input sequences. It is then processed by two parallel branches: a CNN module that captures local n-gram features and a BiLSTM+Attention module that models long-range dependencies. The feature representations from both branches are fused and passed through fully connected layers for the final classification. In the following subsections, we describe each component of the model in detail, beginning with the BERT layer.

\subsubsection{\textbf{Transformer-based Contextual Embedding}}
The contextual embedding layer is based on the pretrained BERT model. Unlike traditional models that use only left context, BERT uses the right context as well for each token, enabling it to assign different embeddings based on each token's context in the text. These embeddings are contextually generated, providing more semantic information than standard static word embeddings. The generated embeddings are then passed to the CNN and BiLSTM branches, where local semantic patterns and long-range contextual dependencies are learned simultaneously. This combination allows the proposed framework to learn more discriminative features while performing both the tasks of headline classification and sentiment analysis. The input configuration of the BERT layer is summarized in Table~\ref{tab:input_config}.

\begin{itemize}
\item\textbf{Input Embeddings:}
Before passing the tokenized text to the BERT encoder, each token is transformed into a dense vector representation that combines three components: token embeddings, positional embeddings, and segment embeddings. For a given token at position $t \in {1,…,T}$ in sample b, the input embedding is defined as\cite{kim2025bert}:
$$h_{b,t}^{(0)} = e_{b,t}(tok)+e_t(pos)+e_{b,t}(seg) \in R^{d_{\text{bert}}}$$
Now for sample b, the input representation is obtained by stacking all of the token embeddings for a sequence.
$$H_b^{(0)} =[h_{b,1}^{(0)}, h_{b,2}^{(0)}, …, h_{b,T}^{(0)}] \in R^{T\times d_{\text{bert}}}$$

\item\textbf{Transformer Encoder:}
The transformer encoder in BERT consists of L stacked layers, each combining multi-head self-attention and a position-wise feed-forward network, with residual connections and layer normalization. This enables the model to capture long-range dependencies and contextual information for each token.

At layer $\ell$, the output of the encoder can be represented compactly as\cite{aurpa2024deep}:\begin{equation*}
\begin{split}
H^{(\ell)} &=
\text{LayerNorm}\Big(
    \text{FFN}\big( \text{MHA}( H^{(\ell-1)}, M ) \big)
    + H^{(\ell-1)}
\Big) \\
&\in \mathbb{R}^{B \times T \times d_{\text{bert}}}
\end{split}
\end{equation*} Where $\text{MHA}(\cdot, M)$ is the masked multi-head self-attention, which computes contextualized token representations while ignoring padded positions indicated by $M. \text{FFN}(\cdot)$ is the position-wise feed-forward network, applied independently to each token embedding. After L layers, the final contextual embeddings are:
$$H = H^{(L)} \in R^{B \times T \times d_{\text{bert}}}$$ These embeddings capture both local token-level patterns and global sequence-level dependencies, and can be used directly for downstream tasks such as classification.

\item\textbf{Dropout Layer:}
A dropout layer is applied immediately after the BERT embeddings to reduce overfitting and improve generalization. Specifically, a dropout rate of 0.35 is used: $$H_{drop} = Dropout(0.35)(H)$$ Here, H is the BERT output $H \in R^{B \times T \times d_{\text{bert}}}$.
\end{itemize}

The $H_{drop}$ is passed to the CNN and BiLSTM modules for downstream feature extraction.

\subsubsection{\textbf{Local Feature Extraction:}}
After obtaining contextual embeddings from BERT, it is essential to capture local patterns and n-gram features that may be indicative of specific aspects or sentiments in news headlines. While BERT embeddings encode rich semantic and contextual information, a CNN layer helps extract position-invariant local features from these embeddings. These are crucial for identifying short-range lexical patterns that can influence classification. We have also employed a 1D convolutional branch with multiple kernel sizes to enhance . This allows the model to capture local patterns of different lengths across the token sequence  $H_{drop} \in R^{B \times T \times d_{\text{bert}}}$.

\begin{itemize}
\item\textbf{Multi-kernel convolution:}
CNN branch detects local semantic patterns on BERT embeddings using 128 convolutional filters that have various kernel sizes, corresponding to various n-gram-length semantic patterns. ReLU activation function is added to enhance feature learning and introduce nonlinearity. L2 regularization helps prevent overfitting by discouraging model weights from being too large; batch normalization helps stabilize training and speeds up convergence. Additionally, a dropout layer with a drop ratio of 0.35 is added to enhance the model's generalization by randomly turning off neurons during training \cite{han2018new}.         \[Z^{(k)}_{b,t,j} = \sigma\left(\sum_{i=1}^{d_{\text{bert}}} \sum_{m=1}^{k}
W^{(k)}_{i,m,j} \cdot H_{\text{drop}}[b, t+m-1, i] + b^{(k)}_{j}\right)
\] where $W^{(k)} \in \mathbb{R}^{d_{\text{bert}} \times k \times 128}$
are convolutional weights, $b^{(k)} \in \mathbb{R}^{128}$ biases, and $\sigma(\cdot)$ denotes ReLU activation.

Batch normalization and dropout($p=0.35$) are then performed after convolution + activation: \[
H^{(k)} = \text{Dropout}\left(\text{BatchNorm}(Z^{(k)})\right). \]

\item\textbf{Pooling and concatenation:}
Global MaxPooling is used to extract the most notable characteristics from each convolution. It pools across the sequence length while picking the maximum activations for each filter: \[
P^{(k)}_{b,j} = \max_{t} H^{(k)}_{b,t,j}, j \in \{1, \dots, 128\}\] Then each of the branches outputs a pooled feature vector: \[
\quad P^{(k)} \in \mathbb{R}^{B \times 128}
\] These pooled features are then concatenated to form a unified feature vector:\[
C = \text{Concat}\left(P^{(2)}, P^{(3)}, P^{(4)}\right) \in \mathbb{R}^{B \times 384}.
\]

\item\textbf{Fully connected projection with regularization:}
This concatenated vector is then goes through the dense layer followed by batch normalization and dropout. The mapping of C to a 256-D representation by the dense layer by the following:\[U_b = \phi\left(C_b W_{\text{fc}} + b_{\text{fc}}\right), \quad
W_{\text{fc}} \in \mathbb{R}^{384 \times 256}, \; b_{\text{fc}} \in \mathbb{R}^{256},\]
where $\phi(\cdot)$ represents batch normalization and ReLU, followed by dropout, $W^{(k)}$ is onvolutional weights, $b^{(k)}$ is convolutional biases, $W_{\text{fc}}$ is dense projection weights and $b_{\text{fc}}$ is biases. Thus, the CNN representation is: \[
CNN = \text{Dropout}\left(\text{BatchNorm}(U_b)\right) \in \mathbb{R}^{B \times 256}.\]
\end{itemize}

\begin{algorithm}
\caption{Feature Extraction using BERT-CNN-BiLSTM with Attention}
\label{alg:feature_extraction}
\begin{algorithmic}[1]

\Require Input sentence $S$, BERT tokenizer $Tok$, maximum sequence length $L_{\max}$, kernel sizes $K=\{2,3,4\}$
\Ensure Hybrid feature vector $F$

\State $(X,M)\gets Tok(S)$
\State Pad/Truncate $X$ to $L_{\max}$
\State $H\gets \mathrm{BERT}(X,M)$
\State $H_{\text{drop}}\gets\mathrm{Dropout}(H)$

\For{$b=1$ to $B$}

    \Statex \textbf{CNN Branch}

    \For{$k\in\{2,3,4\}$}

        \State $Z_k\gets\mathrm{Conv1D}_k(H_{\text{drop}}[b])$
        \State $Z_k\gets\mathrm{ReLU}(Z_k)$
        \State $Z_k\gets\mathrm{BatchNorm}(Z_k)$
        \State $Z_k\gets\mathrm{Dropout}(Z_k)$
        \State $P_k\gets\mathrm{GlobalMaxPool}(Z_k)$

    \EndFor

    \State $C_b\gets\mathrm{Concat}(P_2,P_3,P_4)$
    \State $C_b\gets\mathrm{Dense}(C_b)$

    \Statex \textbf{BiLSTM + Attention Branch}

    \State $H_b\gets\mathrm{BiLSTM}(H_{\text{drop}}[b])$
    \State $H_b\gets\mathrm{BatchNorm}(H_b)$
    \State $H_b\gets\mathrm{Dropout}(H_b)$

    \For{$t=1$ to $T$}

        \State $u_t=\tanh(W_aH_b(t)+b_a)$
        \State $e_t=u_w^{T}u_t$

    \EndFor

    \State $\alpha=\mathrm{Softmax}(e)$

    \State $A_b=\sum_{t=1}^{T}\alpha_tH_b(t)$

    \Statex \textbf{Feature Fusion}

    \State $F_b=\mathrm{Concat}(C_b,A_b)$
    \State $F_b=\mathrm{BatchNorm}(F_b)$
    \State $F_b=\mathrm{Dropout}(F_b)$

\EndFor

\State \Return $F$

\end{algorithmic}
\end{algorithm}

\begin{algorithm}
\caption{Training and Prediction of the Proposed BERT-CNN-BiLSTM Framework}
\label{alg:hybrid_classifier}
\begin{algorithmic}[1]

\Require Hybrid feature vector $F$, label $y$ (training only), smoothing factor $\epsilon$
\Ensure Prediction $\hat{y}$ or loss $L$

\For{$b=1$ to $B$}

    \Statex \textbf{Classification Head}

    \State $U_b \gets \mathrm{Dense}_{192}(F_b)$
    \State $U_b \gets \mathrm{ReLU}(U_b)$
    \State $U_b \gets \mathrm{BatchNorm}(U_b)$
    \State $U_b \gets \mathrm{Dropout}(U_b)$

    \For{$c=1$ to $C$}
        \State $z_{b,c} \gets (W_oU_b+b_o)_c$
    \EndFor

    \State
    $
    \hat{y}_{b,c}
    =
    \frac{\exp(z_{b,c})}
    {\sum_{j=1}^{C}\exp(z_{b,j})}
    $

\EndFor

\If{Training}

    \For{$b=1$ to $B$}

        \For{$c=1$ to $C$}

            \State
            $
            y^{\text{smooth}}_{b,c}
            =
            (1-\epsilon)y_{b,c}
            +
            \frac{\epsilon}{C}
            $

        \EndFor

    \EndFor

    \Statex \textbf{Cross-Entropy Loss}

    \State
    $
    L=
    -\frac{1}{B}
    \sum_{b=1}^{B}
    \sum_{c=1}^{C}
    y^{\text{smooth}}_{b,c}
    \log(\hat{y}_{b,c})
    $

    \State Update all trainable parameters using the Adam optimizer.

    \State \Return $L$

\Else

    \State \Return $\hat{y}$

\EndIf

\end{algorithmic}
\end{algorithm}

\subsubsection{\textbf{Sequential Feature Extraction}}
While the CNN branch captures local n-gram patterns, it is also important to learn global contextual dependencies across the entire sequence. To achieve this, we use a Bidirectional LSTM (BiLSTM) combined with an attention mechanism. The BiLSTM encodes sequential dependencies in both forward and backwards directions, while attention highlights the most informative tokens for the classification task.

\begin{itemize}

\item \textbf{BiLSTM Layer:} The BiLSTM layer models the long-range contextual information from both directions by taking the contextual embeddings generated by BERT, which are represented as $H_{\text{drop}}\in\mathbb{R}^{B\times T\times d_{\text{bert}}}$. For each time step, the LSTM uses the normal gating mechanism:\[
h_t=\mathrm{LSTM}(x_t,h_{t-1}),
\] where the LSTM cell update the forget, input, output, and cell states using the typical gate operations.

The forward and backward hidden states are obtained as: \[
\overrightarrow{h_t}
=
\mathrm{LSTM}_f(x_t,\overrightarrow{h_{t-1}}),
\qquad
\overleftarrow{h_t}
=
\mathrm{LSTM}_b(x_t,\overleftarrow{h_{t+1}}).
\] Finally, the concatenation of the two hidden states is used to get the final BiLSTM representation: \[
h_t=
[\overrightarrow{h_t};\overleftarrow{h_t}],
\qquad
h_t\in\mathbb{R}^{2d_h}.
\] Each of the hidden representations from all time steps are aggregated as \[
H=
[h_1,h_2,\ldots,h_T],\qquad H\in\mathbb{R}^{B\times T\times2d_h}.
\] Finally, to make the training more stable and to prevent overfitting, batch normalization and dropout are put into place. \[H'=\mathrm{Dropout}(\mathrm{BatchNorm}(H)).\]

\item\textbf{Attention mechanism:} The attention mechanism weights each hidden state from the BiLSTM cell and allows the model to pay attention to the most relevant words within the headline. Let $h'_t$ denote the hidden representation at time step $t$ obtained from $H'$. The attention score is computed as \[
u_t=\tanh(W_ah'_t+b_a),
\qquad
u_t\in\mathbb{R}^{d_a},
\] The parameters $W_a$ and $b_a$ are trainable. The normalized attention weight is then obtained by applying the Softmax function: \[
\alpha_t=
\frac{\exp(u_t^{\top}u_w)}
{\sum_{k=1}^{T}\exp(u_k^{\top}u_w)},
\qquad
\alpha_t\in\mathbb{R},
\] where $u_w\in\mathbb{R}^{d_a}$ is the trainable attention context vector.

Finally, the feature representation is gained by adding up all hidden states with weights defined by the attention: \[
\mathrm{BiLSTM}_{\mathrm{att}}
=
\sum_{t=1}^{T}
\alpha_t h'_t ,
\qquad
\mathrm{BiLSTM}_{\mathrm{att}}
\in
\mathbb{R}^{B\times2d_h}.
\]

\end{itemize}

\subsubsection{\textbf{Feature Fusion}}
The strengths of both branches are complementary in that the local features in CNN cooperate with the global contextual features in BiLSTM+Attention. This fusion enables the model to capture both local lexical patterns and logng range dependencies in a single, unified representation.

\begin{itemize}
\item\textbf{Feature Combination}
The CNN feature vector ($\in R^{B×d_{cnn}}$) and BiLSTM-attention feature vector ($\in R^{B×2d_h}$) are concatenated:

\begin{equation*}
\begin{split}
F &= \text{Dropout}\Big(
        \text{BatchNorm}\big(
            \text{Concat}( \text{CNN}, \text{BiLSTM}_{att})
        \big)
    \Big) \\
&\in \mathbb{R}^{B \times (d_{\text{cnn}} + 2d_h)}
\end{split}
\end{equation*}

Batch normalization stabilizes training, while dropout (p=0.35) improves generalization.

\item\textbf{Fully connected layers}
 The combined feature vector is then passed through a dense layer with 192 units, followed by batch normalization and dropout for additional regularization:

$$x = Dropout( BatchNorm( Dense_{192}( F ) ) )$$

\item\textbf{Output layer}
Finally, classification is performed using a softmax layer over C classes, where C corresponds to the number of target labels (e.g., 4 for aspect classification or 3 for polarity classification):

\[
\hat{y} = \text{Softmax}(W_o x + b_o) \in \mathbb{R}^{B \times C}
\]

Where for each class c $\in$ {1,…,C}:
\[
\hat{y}_{b,c} = \frac{\exp\big( (W_o x_b + b_o)_c \big)}{\sum_{j=1}^{C} \exp\big( (W_o x_b + b_o)_j \big)},
\quad \forall c \in \{1, \dots, C\}
\]

\item\textbf{Optimization and loss function}
The model is trained with the AdamW optimizer (learning rate lr, weight decay = 0.01, gradient clipping at 1.0) to improve stability and generalization. To further reduce overconfidence in predictions, categorical cross-entropy loss with label smoothing ($\in$=0.2) is applied:

\[
L = -\frac{1}{B} \sum_{b=1}^{B} \sum_{c=1}^{C}
\Big[ (1 - \epsilon) y_{b,c} + \frac{\epsilon}{C} \Big]
\log \hat{y}_{b,c}
\]

With label smoothing $(\epsilon)$, the target distribution $y^{smooth}_{b,c}$ is defined as:

\[
y^{smooth}_{b,c} = (1 - \epsilon) \cdot y_{b,c} + \frac{\epsilon}{C}
\]
\end{itemize}

Algorithms \ref{alg:feature_extraction} and \ref{alg:hybrid_classifier} outline the entire workflow of the proposed BERT-CNN-BiLSTM framework. The feature extraction is performed using Algorithm 1. The pretrained BERT model is first used to tokenize each input headline and then to generate contextual embeddings for the headline. CNN branch extracts local semantic features by a set of convolution kernels, and the BiLSTM branch learns long-range contextual dependencies. An attention mechanism then determines which of the words are most informative and gives more weight to those words. The features are mixed together from both branches into a single feature representation. The last classification is done by fully connected layers with the Softmax function in Algorithm 2. Label smoothing is used during the training process to prevent overconfidence and to better generalize the model. The model parameters are optimized with the Adam method, and the trained model directly predicts the headline category or sentiment class for the inference.

\subsection{\textbf{Performance Analysis}}
We have evaluated the performance of the models using multiple metrics, including loss, accuracy, precision, recall, and f1-score. Moreover, the model performance was further examined and analyzed through train-test-validation accuracy and loss curves, and XAI-LIME for visualization. The class imbalance was resolved by undersampling, oversampling, and back translation-based augmentation methods in both headline category and sentiment-based datasets. The performance of balanced datasets was verified again using k-fold cross-validation. The results of the analysis are compared thoroughly, as discussed in the next section. It is worth noting that the BAN-ABSA dataset has not been used before. No prior implementation code is available online. Therefore, we have used this dataset for the first time, and the architecture was developed through extensive experimentation and insights from previous studies.

\section{Result Analysis}
\label{sec:4}
In this section, we present experimental results for Bangla news headline classification and sentiment analysis. The BAN-ABSA dataset is highly imbalanced, which can impact model learning and generalization. This problem was solved by two strategies of training. The first strategy only employs oversampling on the training set, whilst the latter only employs back translation-based augmentation on the training set. In both strategies, the validation and test sets are not changed to avoid data leakage and to obtain an unbiased assessment. Transformer-based, deep learning, and hybrid models are compared to the proposed framework based on the accuracy, precision, recall, and F1-score.

\begin{table*}[]
\centering
\resizebox{\textwidth}{!}{\begin{tabular}{|l|ccc|ccc|}
\hline
\multirow{2}{*}{\textbf{Model}}
& \multicolumn{3}{c|}{\textbf{Headline Imbalanced Dataset}}
& \multicolumn{3}{c|}{\textbf{Sentiment Imbalanced Dataset}} \\ \cline{2-7}
& \textbf{Train Acc.} & \textbf{Val. Acc.} & \textbf{Test Acc.}
& \textbf{Train Acc.} & \textbf{Val. Acc.} & \textbf{Test Acc.} \\ \hline

CNN(Frozen)
& 99.93 & 80.93 & 81.33
& 99.54 & 69.49 & 71.65 \\ \hline

BiLSTM(Frozen)
& 99.93 & 76.89 & 77.36
& 99.36 & 63.92 & 67.61 \\ \hline

BERT(Frozen)
& 77.38 & 73.15 & 73.86
& 92.47 & 70.86 & 72.28 \\ \hline

BERT + CNN(Frozen)
& 99.93 & 76.11 & 78.80
& 94.48 & 77.29 & 78.49 \\ \hline

BERT + BiLSTM(Frozen)
& 95.56 & 73.52 & 74.21
& 91.15 & 75.67 & 74.58 \\ \hline

BERT + CNN + BiLSTM(Frozen)
& 93.12 & 74.78 & 75.83
& 95.18 & 77.44 & 79.53 \\ \hline
\end{tabular}}
\caption{Performance Analysis of different of different frozen models on the Headline and Sentiment imbalanced datasets}
\label{tab:frozen_model_comparison}
\end{table*}

\begin{table*}[h!]
\centering
\resizebox{\textwidth}{!}{\begin{tabular}{|l|ccc|ccc|}
\hline
\multirow{2}{*}{\textbf{Model}}
& \multicolumn{3}{c|}{\textbf{Headline Imbalanced Dataset}}
& \multicolumn{3}{c|}{\textbf{Sentiment Imbalanced Dataset}} \\ \cline{2-7}
& \textbf{Train Acc.} & \textbf{Val. Acc.} & \textbf{Test Acc.}
& \textbf{Train Acc.} & \textbf{Val. Acc.} & \textbf{Test Acc.} \\ \hline

CNN & 75.21 & 62.13 & 69.94 & 78.22 & 64.87 & 63.54 \\ \hline

LSTM & 69.84 & 59.47 & 62.40 & 52.59 & 52.52 & 52.57 \\ \hline

BiLSTM & 65.40 & 62.12 & 61.71 & 52.59 & 52.52 & 52.57 \\ \hline

GRU & 72.78 & 52.78 & 55.08 & 52.98 & 50.75 & 51.54 \\ \hline

BiGRU & 78.70 & 60.05 & 60.00 & 52.98 & 50.75 & 51.54 \\ \hline

BERT & 80.72 & 72.55 & 75.50 & 70.20 & 61.25 & 60.14 \\ \hline

BERT + BiLSTM
& 71.45 & 72.20 & 68.23
& 64.65 & 65.10 & 64.80 \\ \hline

BERT + CNN + BiLSTM (No Attention)
& 84.25 & 76.20 & 76.91
& 70.15 & 67.51 & 65.14 \\ \hline
\rowcolor{lightbluebg}
\textbf{Proposed} & \textbf{85.00} & \textbf{77.92} & \textbf{81.37} & \textbf{70.22} & \textbf{67.62} & \textbf{64.46} \\ \hline
\end{tabular}}
\caption{Performance Analysis of different of different finetuned models on the Headline and Sentiment imbalanced datasets}
\label{tab:model_comparison}
\end{table*}

\begin{table*}[h!]
\centering
\small
\resizebox{\textwidth}{!}{\begin{tabular}{|l|c|c|c|c|c|c|}
\hline
\textbf{Dataset} & \textbf{Train (\%)} & \textbf{Val (\%)} & \textbf{Test (\%)} & \textbf{Precision (\%)} & \textbf{Recall (\%)} & \textbf{F1 (\%)} \\ \hline
\rowcolor{lightbluebg}\textbf{Headline – Imb} & 85.00 & 77.92 & 81.37 & 82.49 & 81.37 & 81.54 \\ \hline
\textbf{Headline – UnS} & 79.55 & 76.89 & 75.43 & 79.39 & 75.43 & 75.90 \\ \hline
\textbf{Headline – OvS} & 82.12 & 75.40 & 75.31 & 78.49 & 75.31 & 75.70 \\ \hline
\textbf{Headline – Aug} & 80.48 & 77.46 & 78.63 & \textbf{84.05} & 76.96 & 79.11 \\ \hline
\multicolumn{7}{|c|}{} \\[-1.5ex]
\hline
\textbf{Sentiment – Imb} & 70.22 & 67.62 & 64.46 & 64.70 & 64.46 & \textbf{64.19} \\ \hline
\textbf{Sentiment – UnS} & 62.66 & 60.64 & 59.77 & 66.60 & 59.77 & 60.92 \\ \hline
\textbf{Sentiment – OvS} & \textbf{77.13} & 62.24 & 61.94 & 65.15 & 61.94 & 62.98 \\ \hline
\rowcolor{lightbluebg}\textbf{Sentiment – Aug} & 75.00 & 67.62 & \textbf{66.40} & 66.98 & 66.40 & 63.86 \\ \hline

\end{tabular}
}
\caption{Performance of the proposed BERT--CNN--BiLSTM model under different train-only balancing strategies.}
\label{tab:bert_cnn_bilstm_results}
\end{table*}

\begin{table*}[h!]
\centering
\resizebox{\textwidth}{!}{%
\begin{tabular}{|c|cccccc|cccccc|}
\hline
\multirow{2}{*}{\textbf{Fold}} 
& \multicolumn{6}{c|}{\textbf{Headline – Imbalanced Dataset}} 
& \multicolumn{6}{c|}{\textbf{Sentiment – Augmented Dataset}} \\ \cline{2-13}
& Train & Val & Test & P & R & F1 
& Train & Val & Test & P & R & F1 \\ \hline

1 & 83.62 & 76.48 & 78.06 & 82.36 & 78.06 & 78.44 
  & 91.07 & 64.04 & 68.10 & 63.10 & 60.36 & 61.35 \\ \hline

2 & 81.48 & 78.06 & 77.49 & 80.74 & 77.49 & 77.88 
  & 87.95 & 65.50 & 67.56 & 62.79 & 59.03 & 60.22 \\ \hline

3 & 85.97 & 77.61 & 77.37 & 80.75 & 77.37 & 77.68 
  & 84.59 & 67.62 & 68.59 & 62.92 & 60.08 & 60.90 \\ \hline

4 & 82.43 & 76.11 & 76.91 & 80.01 & 76.91 & 77.21 
  & 92.95 & 66.76 & 68.42 & 62.68 & 60.23 & 61.06 \\ \hline

5 & 81.77 & 76.04 & 76.23 & 80.60 & 76.23 & 76.42 
  & 84.88 & 67.51 & 68.25 & 62.99 & 58.22 & 59.44 \\ \hline

\rowcolor{lightbluebg}
\textbf{Avg} 
  & 83.05 & 76.86 & 77.61 & 80.89 & 77.61 & 77.93
  & 88.29 & 66.29 & 68.18 & 62.90 & 59.58 & 60.59 \\ \hline
\end{tabular}%
}
\caption{K-fold cross-validation results of BERT–CNN–BiLSTM model on Headline and Sentiment Imbalanced datasets}
\label{tab:kfold-tech2-imbalance}
\end{table*}

Tables \ref{tab:bert_cnn_bilstm_results} represent the detailed results of headline and sentiment classification over an imbalanced, undersampled, oversampled dataset,  and augmented dataset . Tables \ref{tab:kfold-tech2-imbalance} represent the stability of the highest result of both headline and sentiment, by using k-fold cross-validation.

\subsection{Performance Analysis under Frozen Models}

We have compared the performance of various architectures based on the frozen BERT on the imbalanced headline and sentiment datasets in Table~\ref{tab:frozen_model_comparison}. For these experiments, the BERT encoder was kept frozen with only the classification layers being trained. The primary aim was to determine whether the extremely imbalanced BAN-ABSA is suffering from overfitting and to validate the need for a more effective training method.

\begin{itemize}

\item \textbf{Headline Classification:} The accuracy of the training results of most frozen models was extremely high, from 93.12\% to 99.93\%. The validation and test accuracies were still quite low, though. The highest test accuracy among these headlines was achieved by Frozen CNN, with 81.33\% and a training accuracy of 99.93\%. The same behavior was seen for the other frozen models. On the large gap to validation data, it is easy to see that the models actually learned what they were supposed to learn rather than features that generalize well. With such a high class imbalance, these results indicate that the headline dataset is indeed overfitting.

\item \textbf{Sentiment Classification:}
A similar trend was observed for sentiment classification. The training accuracy of the frozen models ranged from 91.15\% to 99.54\%, while validation and test accuracy was significantly lower. The highest accuracy was obtained with Frozen BERT-CNN-BiLSTM, with a test accuracy of 79.53\% and 95.18\% during training. This is another large gap in the train-validation split, again indicating strong overfitting. The minority sentiment classes were not learned effectively, so the models work well on the training set, but not so well on the unseen set.

\end{itemize}

It is evident from these experiments that not only is classification accuracy a problem with the BAN-ABSA dataset, but so is generalization. High training accuracy does not necessarily mean the model is reliable. The models are highly prone to overfitting and exhibit poor generalization due to severe class imbalance. Hence, rather than improving training accuracy, this work aims to reduce overfitting. The proposed fine-tuned attention-based BERT-CNN-BiLSTM framework, together with train-only oversampling and back-translation augmentation, allows the model to learn more generalized features while keeping the validation and test sets unchanged. These training methods will remove data leakage and will give an unbiased and fair evaluation.

\subsection{Model Selection and Ablation Study}
Table~\ref{tab:model_comparison} presents the comparison between the proposed BERT-CNN-BiLSTM model and traditional deep learning models, BERT only, and hybrid models on the imbalanced headline and sentiment datasets. The goal is to determine the best possible architecture for the remaining experiments.

\begin{itemize}

\item \textbf{Headline Classification:}
Conventional deep learning models achieved the best test accuracy (69.94\%), whereas a standalone BERT model with contextual text representations attained 75.50\%. Sequential learning alone was not enough, as the BERT-BiLSTM model achieved a test accuracy of 68.23\%. The test accuracy of the hybrid BERT-CNN-BiLSTM model without attention was 76.91\%. The proposed attention-based BERT-CNN-BiLSTM model achieved an accuracy of 81.37\%, which proved that the attention mechanism is useful for focusing the model on the most informative contextual features.

\item \textbf{Sentiment Classification:}
The baseline deep learning models achieved lower performance, and standalone BERT achieved 60.14\% test accuracy. With the BERT-BiLSTM model, the result was further increased to 64.80\%. The best sentiment test accuracy achieved by the BERT-CNN-BiLSTM model without attention was 65.14\%, while the proposed model achieved 64.46\%. The small performance difference indicates that attention has only a limited impact on sentiment classification.

\end{itemize}

The results show that none of the models achieved the best performance on both tasks. The model without attention achieved slightly higher results for sentiment classification and lower results for headline classification, from 81.37\% to 76.91\%. The idea is that BERT is complemented by CNN, BiLSTM, and attention, leveraging their complementary strengths to overcome the limitations of one another. A standalone CNN can capture local features in the form of n-grams, but cannot model the long-range context. BiLSTM is not as effective at extracting local patterns, but it is effective at learning sequential dependencies. While BERT has the ability to capture rich contextual embeddings, it does not explicitly learn task-specific local and sequential features. The attention layer also allows the model to focus on the most informative words while giving less important ones less impact. These complementary components yield a richer feature representation than either the individual components or a simpler hybrid architecture. The proposed attention-based BERT-CNN-BiLSTM is thus capable of achieving the best overall performance on headline classification and sentiment detection. It was based on a single common architecture and was adopted for all later experiments.

\subsection{Performance Analysis Under Different Training Strategies}
Table~\ref{tab:bert_cnn_bilstm_results} presents the performance of the proposed BERT-CNN-BiLSTM model under four training strategies: the original imbalanced dataset, train-undersampled dataset, train-oversampled dataset, and train-augmented dataset. The validation and test sets were not altered to ensure a fair and unbiased evaluation of the training set; only the training set was modified for the balancing strategies.

\begin{itemize}

\item \textbf{Imbalanced Dataset:}
The imbalanced data set that was originally used worked best for the headline classification, with an F1-score of 81.54\% and an accuracy of 81.37\% on the test set. Preserving the semantic diversity of headlines in the natural data distribution enabled the model to learn more representative features. The model test accuracy for sentiment classification is 64.46\%, while the F1 score is 64.19\%. Minority classes were still underrepresented overall, but the results give a baseline to compare to the balancing strategies.

\item \textbf{Train-Undersampled Dataset:}
The train-only undersampling method had the poorest performance on both tasks. The accuracy and the F1-score for the headline and the sentiment datasets were 75.43\% and 75.90\%, respectively, while the accuracy and the F1-score for the sentiment were 59.77\% and 60.92\%, respectively. The elimination of majority-class samples resulted in less useful information for training and reduced semantic diversity, resulting in poorer feature learning and reduced generalization.

\item \textbf{Train-Oversampled Dataset:}
Oversampling to train only gave a better representation of the minority class without changing the validation and test sets. The accuracy of the headline dataset was 75.31\% and F1 score 75.70\% on the test set, and the accuracy of the sentiment dataset was 61.94\% and F1 score 62.98\% on the test set. While the number of training samples was increased with oversampling, duplicated samples incorporated a limited amount of semantic information marginal improvement over undersampling.

\item \textbf{Train-Augmented Dataset:}
The augmented training samples by back translation were semantically similar to the original samples with the same labels. The headline dataset had the highest test accuracy of 78.63\%, and best precision of 84.05\%, while the sentiment dataset had the best test accuracy of 66.40\%. Finally, these results show that the augmented data can lead to higher lexical diversity and better learning of features, but without any evaluation bias. The results of the two tasks are comparable, indicating that train-only augmentation is most effective for minimizing the impact of class imbalance and enhancing model generalization.

\end{itemize}

\subsection{K-fold Cross-Validation}

The 5-fold cross-validation result of the proposed BERT-CNN-BiLSTM model for the best training strategies for each task is shown in Table~\ref{tab:kfold-tech2-imbalance}. Based on the experimental results, the original imbalanced dataset was chosen for headline classification,, and the train-augmented dataset for sentiment classification.

\begin{itemize}

\item \textbf{Headline Classification:}
The proposed model was stable over the 5 folds on the imbalanced dataset. The average training, validation, and test accuracies were 83.05\%, 76.86\%, and 77.61\%, respectively, with an average Precision of 80.89\%, Recall of 77.61\%, and F1-score of 77.93\%. There is a negligible difference across a few-fold variation; thus, it generalizes well to the data distribution encountered in nature and remains consistent in delivering headline classification performance.

\item \textbf{Sentiment Classification:}
The train-augmented dataset also showed consistent results across all 5 folds. The proposed model achieved a Precision of 62.90\%, a Recall of 59.58\%, an F1 of 60.59\%, and an average training, validation, and test accuracy of 88.29\%, 66.29\%, and 68.18\%, respectively. The stable generalization across data splits and the robustness of the model are both demonstrated in the consistent performance of the model, which was enhanced by back translation.

\end{itemize}

The results of cross-validation confirm that the proposed framework keeps the performance unchanged across different training and testing partitions. Again, the results validate that the original imbalanced dataset is the best choice for classification of headlines, while train-only back-translation-based augmentation is the most effective choice for sentiment classification. The results also demonstrate that the proposed framework does not cause evaluation bias and that it generalizes well.

\begin{figure*}[!t]
\centering

\subfloat[Headline Classification LIME Explanation]{
    \includegraphics[width=0.48\textwidth]{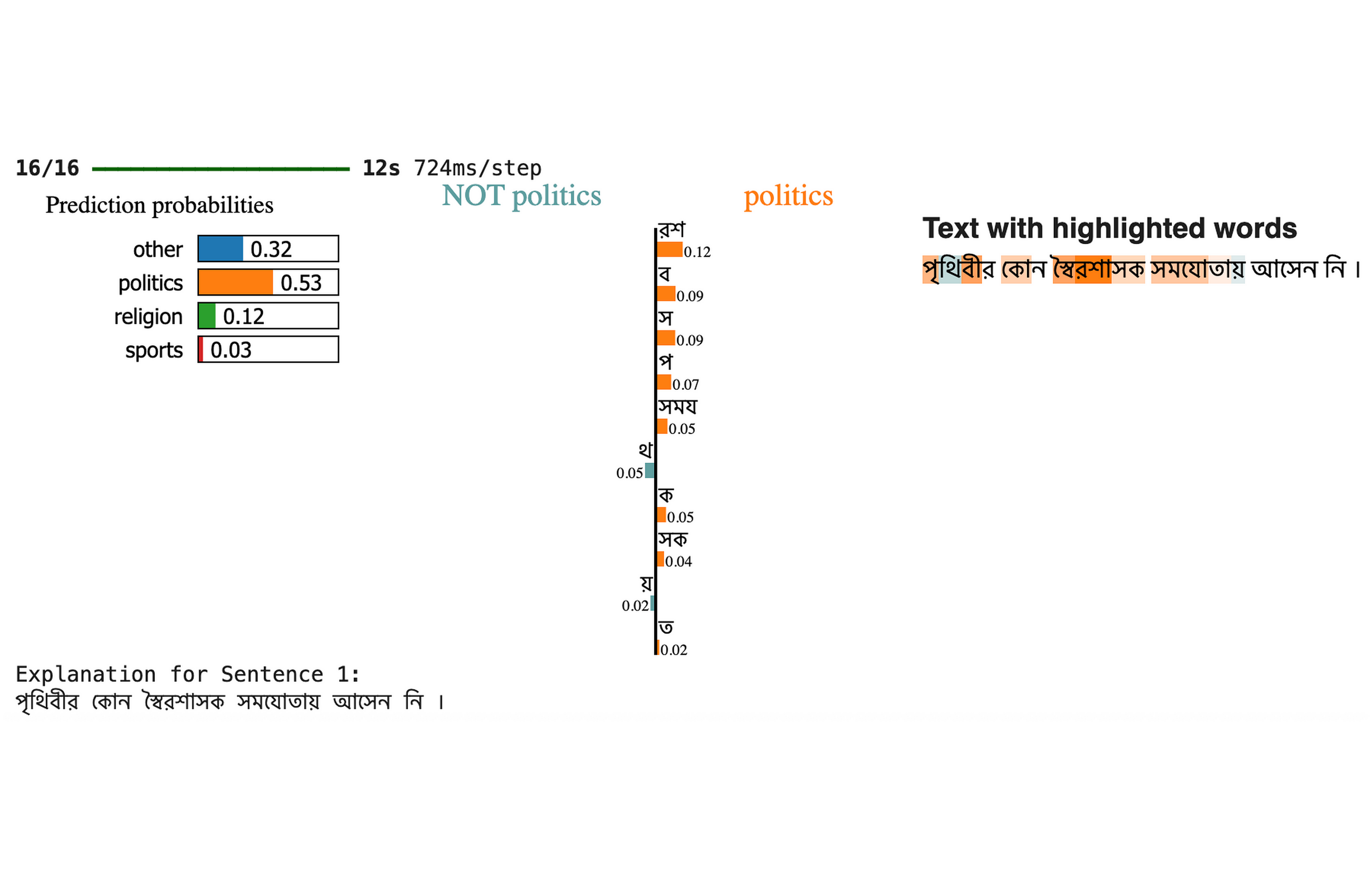}
    \label{fig:lime_headline}
}
\hfill
\subfloat[Sentiment Classification LIME Explanation]{
    \includegraphics[width=0.48\textwidth]{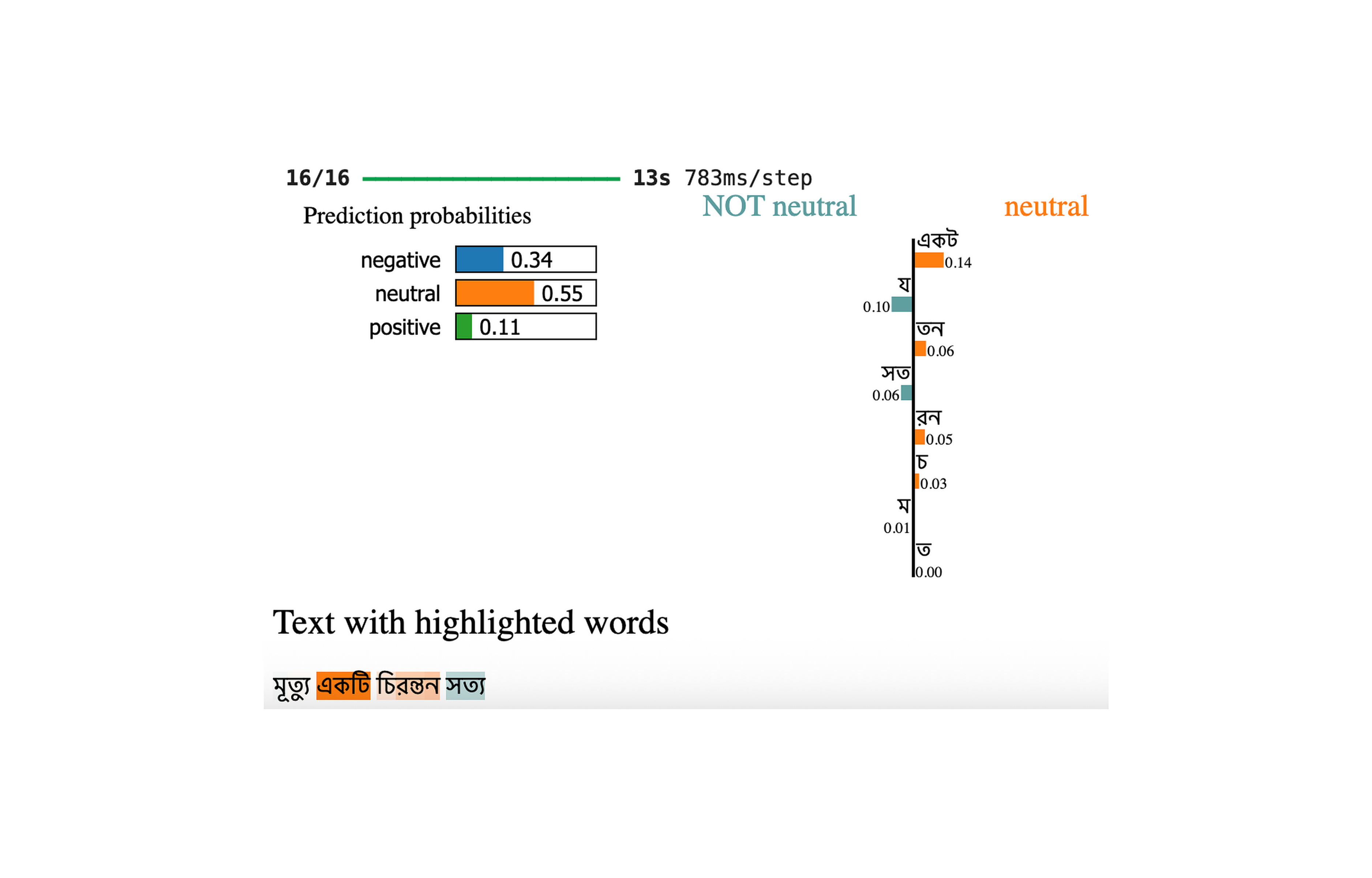}
    \label{fig:lime_sentiment}
}

\caption{LIME explanations of the proposed BERT--CNN--BiLSTM model. (a) headline classification and (b) sentiment classification.}
\label{fig:lime_explanations}
\end{figure*}

\subsection{Model Interpretability Using LIME}

This section presents the interpretability analysis of the proposed model using Local Interpretable Model-agnostic Explanations (LIME).

\begin{itemize}
\item \textbf{Headline Classification:} As shown in the top of Figure~\ref{fig:lime_explanations}, the LIME explanation assigned the Bangla headline to the \textit{Politics} class with a prediction probability of 0.53, while the \textit{Other} class received a probability of 0.32. The words with positive weights (red) strongly supported the \textit{Politics} class, whereas a few words with negative weights (blue) provided weak opposing evidence. The result shows that the proposed model successfully identifies domain-specific political terms for headline classification.

\item \textbf{Sentiment Classification:} Figure~\ref{fig:lime_explanations} (bottom) shows the LIME explanation for sentiment classification. The model predicted the \textit{Neutral} class with a probability of 0.55, while the \textit{Negative} and \textit{Positive} classes received probabilities of 0.34 and 0.11, respectively. The highlighted words contributed strongly to the \textit{Neutral} prediction, whereas a few tokens provided weak support for other sentiment classes. These results indicate that the proposed model relies on contextual semantic information rather than isolated emotional words for sentiment prediction.

\begin{figure}[]
\centering
\includegraphics[width=\columnwidth]{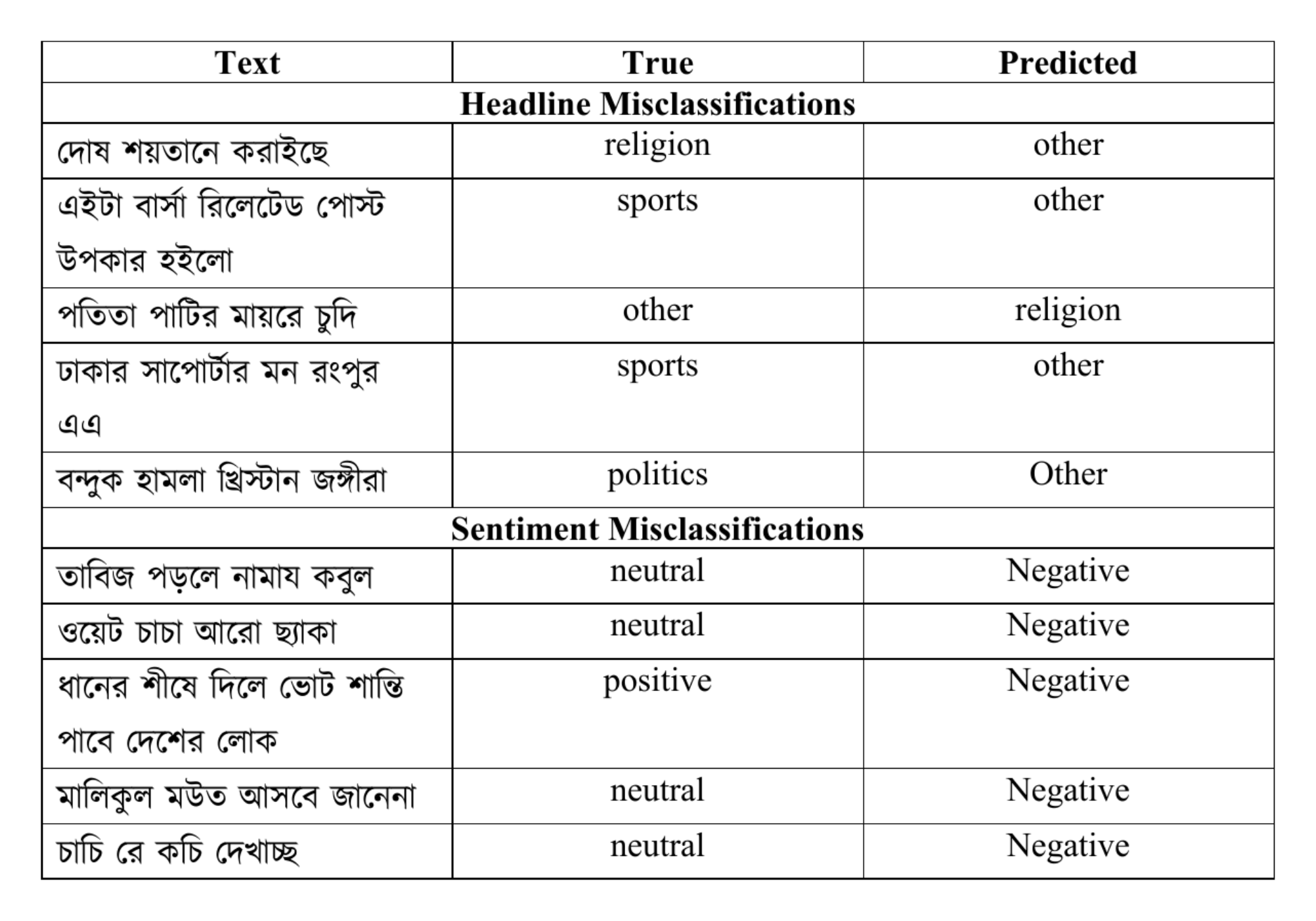}
\caption{Representative headline and sentiment misclassification samples}
\label{fig:misclassified_examples}
\end{figure}
\end{itemize}

\begin{table}[!ht]
\centering
\small
\begin{tabular}{|l|l|}
\hline
\textbf{Headline Classification} & \textbf{Sentiment Classification} \\ \hline
Politics $\rightarrow$ Other (71) &
Neutral $\rightarrow$ Negative (110) \\

Sports $\rightarrow$ Other (51) &
Positive $\rightarrow$ Negative (93) \\

Religion $\rightarrow$ Other (36) &
Positive $\rightarrow$ Neutral (30) \\

Other $\rightarrow$ Religion (32) &
Negative $\rightarrow$ Neutral (18) \\

Other $\rightarrow$ Politics (28) &
Negative $\rightarrow$ Positive (14) \\ \hline
\end{tabular}
\caption{Most frequent misclassification pairs.}
\label{tab:confused_pairs}
\end{table}

\begin{table}[!ht]
\centering
\small
\resizebox{1.0\columnwidth}{!}{\begin{tabular}{|l|c|c|}
\hline
\textbf{Metric} & \textbf{Headline} & \textbf{Sentiment} \\ \hline
FLOPs & 55,064,479,196 & 55,064,478,806 \\ \hline
GFLOPs & 55.064 & 55.064 \\ \hline
Training Time (s) & 1973.30 & 5449.79 \\ \hline
Avg. Time/Epoch (s) & 219.26 & 217.99 \\ \hline
Inference Latency (ms/sample) & 143.876 & 145.147 \\ \hline
\end{tabular}}
\caption{Computational complexity of the proposed BERT--CNN--BiLSTM framework.}
\label{tab:complexity_analysis}
\end{table}

\begin{table}[h!]
\centering
\resizebox{1.0\columnwidth}{!}{\begin{tabular}{|c|cccccc|}
\hline
\textbf{Dataset} & \textbf{Train} & \textbf{Val} & \textbf{Test} & \textbf{P} & \textbf{R} & \textbf{F1} \\ \hline

Potrika Dataset
& 85.42 & 84.00 & 84.34 & 84.62 & 84.34 & 84.30 \\ \hline

Advanced Sentiment
& 77.56 & 71.73 & 71.52 & 74.72 & 71.52 & 71.52 \\ \hline

\end{tabular}}
\caption{Performance of BERT–CNN–BiLSTM on External Datasets}
\label{tab:external-datasets-tech2}
\end{table}

\begin{table}[t]
\centering
\resizebox{\columnwidth}{!}{\begin{tabular}{|l|cccc|cccc|}
\hline
\multirow{2}{*}{\textbf{Method}}
& \multicolumn{4}{c|}{\textbf{Headline Classification}}
& \multicolumn{4}{c|}{\textbf{Sentiment Classification}} \\ \cline{2-9}

& \textbf{Acc.} & \textbf{P} & \textbf{R} & \textbf{F1}
& \textbf{Acc.} & \textbf{P} & \textbf{R} & \textbf{F1} \\ \hline

BAN-ABSA(CNN/BiLSTM)
& 78.75 & 80.46 & 78.33 & 79.38
& 71.08 & 62.49 & 62.11 & 62.30 \\ \hline

\rowcolor{lightbluebg}
\textbf{BERT + CNN (Frozen)}
& \textbf{78.80} & \textbf{79.64} & \textbf{78.59} & \textbf{79.07}
& \textbf{78.49} & \textbf{75.78} & \textbf{70.98} & \textbf{72.28} \\ \hline

\end{tabular}
}
\caption{Comparison with the baseline results reported on the BAN-ABSA dataset.}
\label{tab:comparison_baseline}
\end{table}

\subsection{Error Analysis and Misclassification Study}
The most common pairs of confusions of errors are summarized in Table~\ref{tab:confused_pairs}. The highest error rates in headline classification were on the Sports category, and the largest number of errors were classified as other. The neutral class was the most misclassified in sentiment classification as most samples were classified as negative. The errors are primarily related to the overlapping of meaning and the non-specific expression of context. Even though the cases are hard, the proposed BERT--CNN--BiLSTM model outperforms stably and has decent generalization on both tasks. Some misclassified samples are also shown in Fig-\ref{fig:misclassified_examples}

\subsection{Computational Complexity}
The complexity analysis of the proposed framework is summarized in Table~\ref{tab:complexity_analysis}. The two tasks take about 55.06 GFLOPs.The average epoch training time is approximately 218 seconds, and the inference latency is approximately 144 milliseconds per sample, showing real-world computational efficiency for deployment.

\subsection{External Dataset Validation}

\begin{itemize}

\item \textbf{Potrika Dataset -- Headline Classification:}
The proposed BERT-CNN-BiLSTM model was evaluated on the external Potrika dataset, where 24,000 balanced headlines (3,000 headlines per class) were sampled from a large corpus of 665K news articles \cite{ahmad2022potrika}. As shown in Table~\ref{tab:external-datasets-tech2}, the model achieved 84.34\% test accuracy, 84.62\% precision, 84.34\% recall, and an F1-score of 84.30\%. These results demonstrate that the proposed model generalizes well to large-scale Bangla news datasets and effectively classifies diverse headline categories.

\item \textbf{Advanced Sentiment Dataset -- Sentiment Classification:}
The proposed model was also evaluated on the Advanced Sentiment dataset containing 8,738 negative, 7,328 strongly negative, 7,697 neutral, 6,246 positive, and 4,803 strongly positive samples \cite{tripto2018detecting}. The model achieved 73.05\% test accuracy, 76.28\% precision, 73.05\% recall, and a 73.39\% F1-score without using oversampling. These results demonstrate that the proposed hybrid framework generalizes well and effectively captures sentiment patterns across multiple sentiment classes.
\end{itemize}

\subsection{Comparison with Existing Work}

The BAN-ABSA dataset was initially published with two baseline models (CNN and BiLSTM) for two tasks, headline classification and sentiment classification. The baseline study achieved 78.75\% accuracy in headline classification and 71.08\% in sentiment classification and recommended introducing attention-based models for potential future improvements \cite{ahmed2021ban}. Table~\ref{tab:comparison_baseline} presents a direct comparison between the original BAN-ABSA baseline and our proposed framework, highlighting the performance improvements achieved on the same dataset.

In contrast to earlier research, this paper introduces a joint BERT--CNN--BiLSTM model for determining a headline's classification category and sentiment. The same architecture is fine-tuned for both tasks instead of designing separate models. The BAN-ABSA data set is extremely imbalanced, which this study addresses by oversampling and back-translating only the training set, rather than the validation and test sets. This approach will help avoid data leakage, minimize overfitting, and provide a fair evaluation. The proposed framework has been tested on extensive datasets, and ablation results reveal that it provides a more robust and generalizable solution for Bangla text classification. Some frozen BERT models achieved higher training accuracy but suffered from severe overfitting, as evidenced by the significant difference between training and validation performance. The proposed fine-tuned BERT--CNN--BiLSTM model achieves a better trade-off between training and test performance, thereby improving the model's transferability between the two tasks.

\section{Conclusion}
\label{sec:5}
This paper proposed a single BERT-CNN-BiLSTM framework for Bangla headline classification and sentiment analysis. The proposed framework was a train-only approach that used undersampling, oversampling, and back-translation-based augmentation. This study addresses severe class imbalance and overfitting on the BAN-ABSA dataset, while the validation and test sets remained unchanged to ensure a fair and unbiased assessment. The model was tested with four different train settings, i.e., imbalanced, train-undersampled, train-oversampled, and train-augmented datasets. Experimental results revealed that the original imbalanced dataset performed best for headline classification, while the train-augmented one performed best for sentiment classification. The proposed framework was further validated using five-fold cross-validation, external dataset validation, and LIME-based interpretability analysis, which demonstrated its robustness and generalization capability. The proposed framework creates a good reference point for the BAN-ABSA dataset and offers an effective approach for dealing with Bangla text classification with class imbalance. Future research aims to study code-mixed Bangla-English text, increase the size of benchmark sets, and fine-tune modern large language models (LLMs) to further enhance classification performance and generalization ability.

\bibliographystyle{IEEEtran}

\end{document}